\documentclass[a4paper,11pt]{article}

\usepackage{fourier}
\usepackage{color}
 \usepackage{graphicx}
\usepackage{url}
\usepackage[affil-it]{authblk}
\usepackage{amsmath}
\usepackage{wrapfig}

\usepackage[T1]{fontenc}
\usepackage{times}
\usepackage[hidelinks]{hyperref}

\begin{document}

\title{MapsTP: HD Map Images Based Multimodal Trajectory Prediction for Automated Vehicles}
\author{Sushil Sharma$^{1,2}$, Arindam Das$^{1}$, Ganesh Sistu$^{1}$,  Mark Halton$^{1}$ and Ciar\'{a}n Eising$^{1,2}$ }
\affil{$^{1}$Department of Electronic \& Computer Engineering, University of Limerick, Ireland\\ $^{2}$SFI CRT Foundations in Data Science, University of Limerick, Ireland}
\affil{$^{1}$firstname.lastname@ul.ie}

\affil{\bf This paper is a preprint of a paper submitted to the 26th Irish Machine Vision and Image Processing
Conference (IMVIP 2024). If accepted, the copy of record will be available at IET Digital Library.}
%\author{Anonymous Submission}
%\affil{Anonymous Affiliation}
\date{}
\maketitle
\thispagestyle{empty}

\begin{abstract}
Predicting ego vehicle trajectories remains a critical challenge, especially in urban and dense areas due to the unpredictable behaviours of other vehicles and pedestrians. Multimodal trajectory prediction enhances decision-making by considering multiple possible future trajectories based on diverse sources of environmental data. In this approach, we leverage ResNet-50 to extract image features from high-definition map data and use IMU sensor data to calculate speed, acceleration, and yaw rate. A temporal probabilistic network is employed to compute potential trajectories, selecting the most accurate and highly probable trajectory paths. This method integrates HD map data to improve the robustness and reliability of trajectory predictions for autonomous vehicles.
\end{abstract}

%%%%%%%%%%%%%%%%%%%%%%
\section{Introduction}
\begin{wrapfigure}{r}{0.5\textwidth}
  \vspace{-20pt}
  \begin{center}
    \includegraphics[width=0.48\textwidth, height=0.18\textwidth]{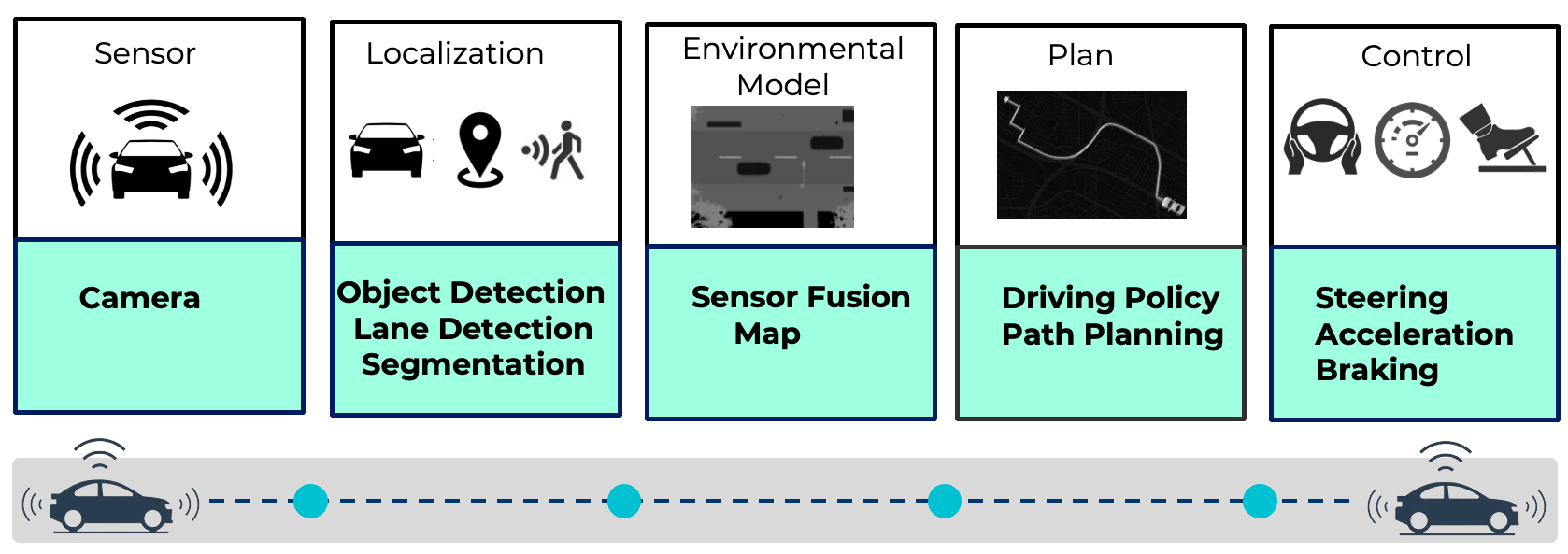}
\end{center}
\vspace{-20pt}
  \caption{Standard Autonomous Driving Pipeline Overview}
  \vspace{-0pt}
  \label{fig:baseline}
\end{wrapfigure}
%%%%%%%%%%%%%%%%%%%%%%
Automated vehicles (AVs) are changing the way we think about transportation, promising to make our roads safer and more efficient \cite{hancock2019future,chehri2019autonomous,Sharma2024BEVSeg2TP}. A key challenge for AVs is predicting the paths of ego vehicle, other vehicles, pedestrians, and cyclists. This is essential to avoid collisions \cite{zhao2017review} and ensure smooth driving \cite{huang2023differentiable}. Recently, researchers have been exploring the use of high-definition (HD) maps combined with various data sources to improve these predictions \cite{elghazaly2023high,bao2023review}. HD maps offer detailed information about the road, including lane markings, traffic signs, and road shapes. This information can help AVs better understand their surroundings and make more accurate decisions. By combining HD maps with trajectory prediction models, recent studies have shown improved accuracy in predicting where other road users will go \cite{guo2023map}. Figure \ref{fig:baseline}
shows the standard end-to-end pipeline containing essential components for standard autonomous driving systems.\\

In this paper, we introduce the model MapsTP, which uses detailed information from HD maps to enhance trajectory prediction. It also considers sensor data to understand the different behaviours of road users like cars, pedestrians, and cyclists. This comprehensive approach is crucial for developing AVs that can navigate urban city environments safely and efficiently. Our model predicts ego vehicle trajectories by combining HD map images and dynamic sensor data. HD maps provide detailed environmental information, while sensors capture the vehicle's state. Features are extracted using ResNet-50 and fed through a Spatio-Temporal Probabilistic Network to estimate multiple potential trajectories, selecting the most likely one. Our method simplifies the model and enhances prediction accuracy, as shown in the results section.\\

In the present study, the scope of this research is to enhance the precise prediction of the ego vehicle trajectory and contrast it with existing benchmarks. The primary achievements of this research include:

\begin{enumerate}
    \item  We developed a model to predict ego vehicle trajectory using HD map data. The detailed environmental features and vehicle states are processed to generate multiple possible trajectories.
    \item  We compared our model with four leading baseline models for trajectory prediction using the nuScenes dataset \cite{caesar2020nuscenes}. Our evaluation showed how our model performs against established methods, demonstrating its effectiveness in predicting vehicle paths.
    
   % \item  Our model uses HD maps to provide better context, resulting in more accurate predictions in complex urban environments. This approach helps the model understand road layouts and traffic patterns more effectively
\end{enumerate}

%These advancements have laid the groundwork for new solutions like MapsTP, which aims to improve AV technology by integrating HD maps and diverse data sources for better trajectory prediction.

%%

\section{Research Background}
Research into predicting the future movement of vehicles has been ongoing for many years. Initial studies \cite{kaempchen2004imm,lytrivis2008cooperative} utilized physics-based kinematic models to estimate future positions, demonstrating good accuracy in short-term trajectory  \cite{sharma2023navigating}. However, these models struggle to accurately predict long-term vehicle behaviour, which is heavily influenced by external factors such as weather conditions, traffic conditions, and the intentions of the drivers such as lane changes, navigation goals \cite{fang2024behavioral,cong2023dacr}. More recent investigations have turned to deep learning techniques, specifically recurrent networks, to enhance prediction accuracy. These methods incorporate historical data about the actors involved and contextual information from neighbouring entities, offering a more nuanced approach to vehicle movement forecasting.\\

Multimodal prediction is a technique used to predict a set of possible paths for each object involved, taking into account its unpredictable behaviour, such as changes in speed and direction. To address the difficulty of predicting multiple outcomes, deep learning techniques are often used. These include generative adversarial networks \cite{goodfellow2014generative}, and Graph neural networks \cite{sharma2024optimizing}. Each of these systems uses unique approaches to solve multimodal trajectory problems. Recent studies by \cite{9562034, yuan2021agent, lee2022muse} and \cite{chen2022scept} are based on the concept of using a conditional variational autoencoder to predict multiple vehicle trajectories efficiently. Despite their advantages, these sampling-based methods lack an explicit way to calculate the probability of each predicted path during the random selection stage. Additionally, with the remarkable progress in computer vision, diffusion models have begun to be used to understand the behaviour of road users and generate various vehicle trajectory predictions. However, these models usually require a lot of time due to the use of sequential sampling procedures. Inspired by these previous methods, we designed our model, MapsTP, which estimates precise trajectory predictions of the ego vehicle using HD map images and a spatio-temporal probabilistic network.

\section{Our Approach}

In this section, we explain how our proposed architecture works and detail each component involved. Our methodology integrates HD maps, feature extraction, and a spatio-temporal probabilistic network to predict vehicle trajectories accurately. 

\begin{figure}[h!]
\centering
\includegraphics[width=\columnwidth]{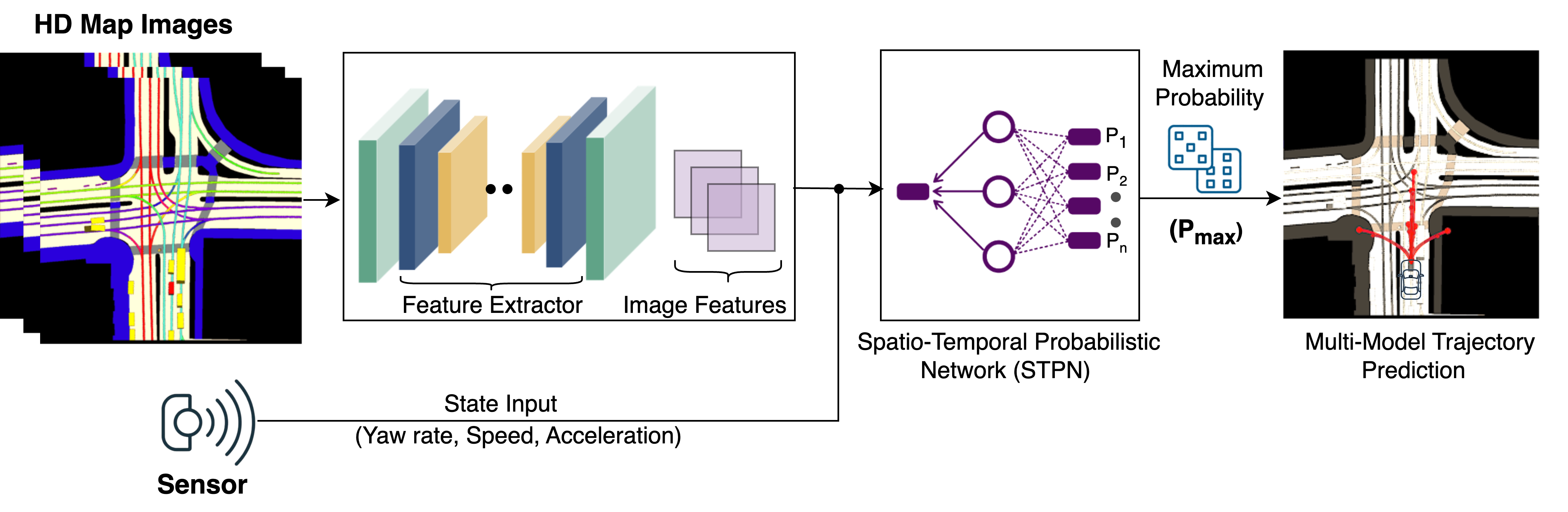}
\caption{\textbf{Our proposed architecture}: HD Map Integration for Vehicle Trajectory Planning. The process begins with the feature extractor analyzing these inputs to create image features. These features are then analyzed by a probabilistic network, which evaluates multiple potential trajectories and assigns probabilities to each.} 
\label{fig:boschVirtualVisor}
\end{figure}

\subsection{HD Maps}
HD maps are essential for providing detailed and accurate information about the environment surrounding the vehicle. These maps include precise lane markings, road boundaries, traffic signs, and other critical features that are necessary for safe navigation. By utilizing HD maps, the system can have a comprehensive understanding of the road structure and layout, which is crucial for accurate trajectory prediction.

\subsection{Feature Extractor}

The feature extractor (ResNet-50) block is responsible for processing HD map data and sensor inputs to generate meaningful features for the prediction model. The sensor inputs typically include yaw rate, speed, and acceleration, providing real-time information about the vehicle's state. The feature extractor converts raw data from the IMU sensor into discrete segments corresponding to vehicle movement over time, highlighting important characteristics such as lane positions, intersections, and road geometry. These extracted features are then fed into the next stage of the system.

\subsection{Spatio-Temporal Probabilistic Network (STPN)}
The STPN is designed to predict possible trajectories based on the extracted features. This network considers the spatial relationships (such as road layout and agents) and temporal dynamics (such as vehicle speed and acceleration) to generate a set of probable trajectories \( P_1, P_2, \ldots, P_n \). Each individual trajectory \( P_i \) is assigned a probability score that indicates the likelihood of it being the correct prediction.

The probabilistic network can be represented by a function \( g \) that maps the feature vectors \( F \) to a set of possible trajectories with associated probabilities:
\[ \{(P_i, \text{prob}(P_i))\} = g(F) \quad \text{for} \quad 1 \leq i \leq N\]
where \( \text{prob}(P_i) \) is the probability of the \( i \)-th trajectory. we typically denote the range of indices over which 
$i$ varies. If there are  $N$ possible trajectories, the index  $i$ ranges from 1 to  $N$

\subsection{Maximum Probability and Output Trajectory}
Once the trajectories are generated, the system identifies the trajectory with the maximum probability. This trajectory represents the most likely path the vehicle will follow. The final step involves projecting this trajectory back onto the HD map, resulting in the output trajectory that guides the vehicle's movement.

The output trajectory ensures that the vehicle follows a safe and efficient path, taking into account the detailed road information from the HD maps and the real-time state inputs.
%% comment line

\section{Implementation Details}

We implemented our method using ResNet-50 \cite{he2016deep} with pre-trained ImageNet \cite{russakovsky2015imagenet} weights as the backbone. High-level features are extracted from the conv5 layer and globally pooled, then combined with an agent state vector that includes velocity, acceleration, and yaw rate as described in \cite{cui2019multimodal}. This model predicts trajectories within a 6-second horizon. Training parameters include a fixed learning rate of $1e-4$, a batch size of $32$, and $50$ epochs. This setup effectively integrates visual and dynamic information to achieve reliable trajectory prediction.

\section{Experimental setup and Evaluations}

In this section, we discuss the dataset utilized in our research and delve into the specifics of the evaluation metrics employed to assess the performance of our baseline models.

\subsection{Nuscenes Dataset}
To evaluate our method, we considered nuScenes dataset \cite{caesar2020nuscenes}, which comprises $1000$ video sequences captured in Boston and Singapore. Each video is 20 seconds in duration and contains 40 frames, resulting in a total of 40,000 samples. The dataset is divided into training, validation, and testing sets, with each set containing 700, 150, and 150 scenes, respectively.

\subsection{Evaluation Metrics}
We calculate evaluation metrics for our model by comparing it with state-of-the-art models, using a comprehensive set of performance metrics to evaluate its effectiveness and robustness, as in previous studies \cite{phanminh2020covernet,chai2019multipath}\\

\paragraph{MinADE${_k}$}: We compute the minimum average and final displacement errors across $k$ predicted trajectories for ego vehicle trajectory prediction.
\begin{equation} \label{eqn:ADE}
\mathbf{MinADE_k} =  \min_{i\in\{1...K\}} \frac{1}{T_f} \sum_{t=1}^{T_f} \left\Vert y_t^{\text{gt}} - y_t^{(i)} \right\Vert_2
\end{equation}
where $y^{gt}$ represents the ground truth position of the object at the final time step T, and $y_{t}^{(i)}$ represents the predicted position of the object at the final time step T for the $i_{th}$ trajectory in the set of k trajectories.

\paragraph{MinFDE${_k}$}: We evaluate the minimum final displacement error across $k$ predicted trajectories for ego vehicle trajectory prediction.
\begin{equation} \label{eqn:FDE}
\mathbf{MinFDE_k} =  \min_{i\in\{1...K\}} \left\Vert y_t^{\text{gt}} - y_t^{(i)} \right\Vert_2
\end{equation}

\paragraph{MissRate$_{k,d}$}: Similar to \cite{messaoud2020trajectory}, to calculate the miss rate, we consider a given distance $d$ and the $k$ most likely predictions generated by the STPN model. If the minimum distance between these predictions and the ground truth is greater than \textit{d}, we count it as a miss.
\begin{equation} \label{eqn:miss}
\mathbf{Miss Rate_{k,\textit{d}}}= 
\begin{cases} 
1 & \text{if} \ \min_{\hat{y} \in P_k} \left( \max_{t=P_t}^{t=F_t} \left\Vert y_t^{\text{gt}} - y_t^{(i)} \right \Vert \right) \geq d \\
0 & \text{otherwise} 
\end{cases}
\end{equation}
Where ${F_t}$ donated the true future time step, whereas ${P_t}$  represented the predicted future time step.

\section{Results}

In this section, we present our analysis, which includes both quantitative and qualitative components. Our quantitative analysis provides a statistical overview of the data, highlighting key trends and patterns, while our qualitative analysis offers in-depth insights and a contextual understanding of the results.

\subsection{Quantitative Analysis}
We carry out a thorough experimental analysis, comparing our model against four benchmark models:\\\cite{phanminh2020covernet}, \cite{cui2019multimodal}, \cite{chai2019multipath}, and \cite{salzmann2021trajectron}. These benchmarks represent the state-of-the-art in trajectory prediction according to recent proposed models. The purpose of this comparison is to determine how well our model performs in predicting trajectories compared to these established models. We seek to establish if our model outperforms or aligns with the capabilities of current benchmarks. Through this evaluation, we aim to identify the model's strengths and weaknesses, which helps us to identify areas for potential improvement. To assess our model's performance on the nuScenes dataset, we examine the generated trajectory sequence \([ y_1, y_2, y_3, \ldots, y_n ]\).\\

We evaluated the performance of the model on the dataset for different values of $K$, where $K$ is set to 5, 10, and 15. To compute MinADE$_{k}$, MinFDE$_{k}$ and MissRate$_{k,d}$  we use equation \ref{eqn:ADE}, \ref{eqn:FDE} and \ref{eqn:miss}. During the training phase, our goal was to minimize the MinADE$_{k}$ on the training set, as outlined in Table \ref{table:1}. Essentially, we sought to lessen the discrepancy between the predicted and actual trajectories by minimizing the smallest difference between them across each of the $K$ time intervals. This strategy enabled us to enhance the precision of our model's predictions and confirm its efficacy on the nuScenes dataset.
\\

\begin{table*}[h!]
\centering

\scalebox{0.66}{
\begin{tabular}{c|cccccccc}
\hline
\textbf{Method} &  MinADE$_{5}$ $\downarrow$ & MinADE$_{10} \downarrow$  & MinADE$_{15} \downarrow$  & MinFDE$_{5}$ $\downarrow$ & MinFDE$_{10}$ $\downarrow$ & MinFDE$_{15}$ $\downarrow$ & MissRate$_{5,2}$ $\downarrow$ & MissRate$_{10,2}$ $\downarrow$\\
  \hline
  \hline

\textit{CoverNet} \cite{phanminh2020covernet} & 2.62 & 1.92 & 1.63 & 11.36 & - & - & 0.76 & 0.64 \\
\textit{Trajectron++ }\cite{salzmann2021trajectron} & 1.88 & 1.51 & - & - & - & - & 0.70 & 0.64 \\ 
\textit{MTP} \cite{cui2019multimodal} & 2.22 & 1.74 & 1.55 & 4.83 & 3.54 & 3.05 & 0.74 & 0.67\\ 
\textit{MultiPath} \cite{chai2019multipath} & 1.78  & 1.55 & 1.52 & 3.62 & 2.93 & 2.89 & 0.78 & 0.76 \\
\textit{MHA-JAM} \cite{messaoud2020trajectory} & 1.85  & 1.24 & \textbf{1.03} & 3.72 & 2.23 & 1.67 & 0.60 & 0.46 \\
\textit{GoalNet} \cite{zhang2020mapadaptive} & 1.80  & 1.64 & 1.52 & 4.65 & 3.83 & - & 0.58 & 0.45 \\ \hline
\textit{\textbf{MapsTP (Ours)}}   & \textbf{1.30} & \textbf{1.12} &  1.11 &
\textbf{3.58} & \textbf{2.19} & \textbf{1.55} & \textbf{0.51} & \textbf{0.42} \\ \hline

\end{tabular}
}
\caption{Comparison of Methods on nuScenes Dataset \cite{caesar2020nuscenes}: \textbf{Minimum Average Prediction Error (MinADE)} and \textbf{Final Displacement Error (MinFDE)} Analysis over a 6-second prediction horizon}\label{table:1}
\end{table*}
\label{sec:results}

\subsection{Qualitative Analysis}

\begin{figure*}[h!]
\centering
\includegraphics[width=0.85\textwidth]{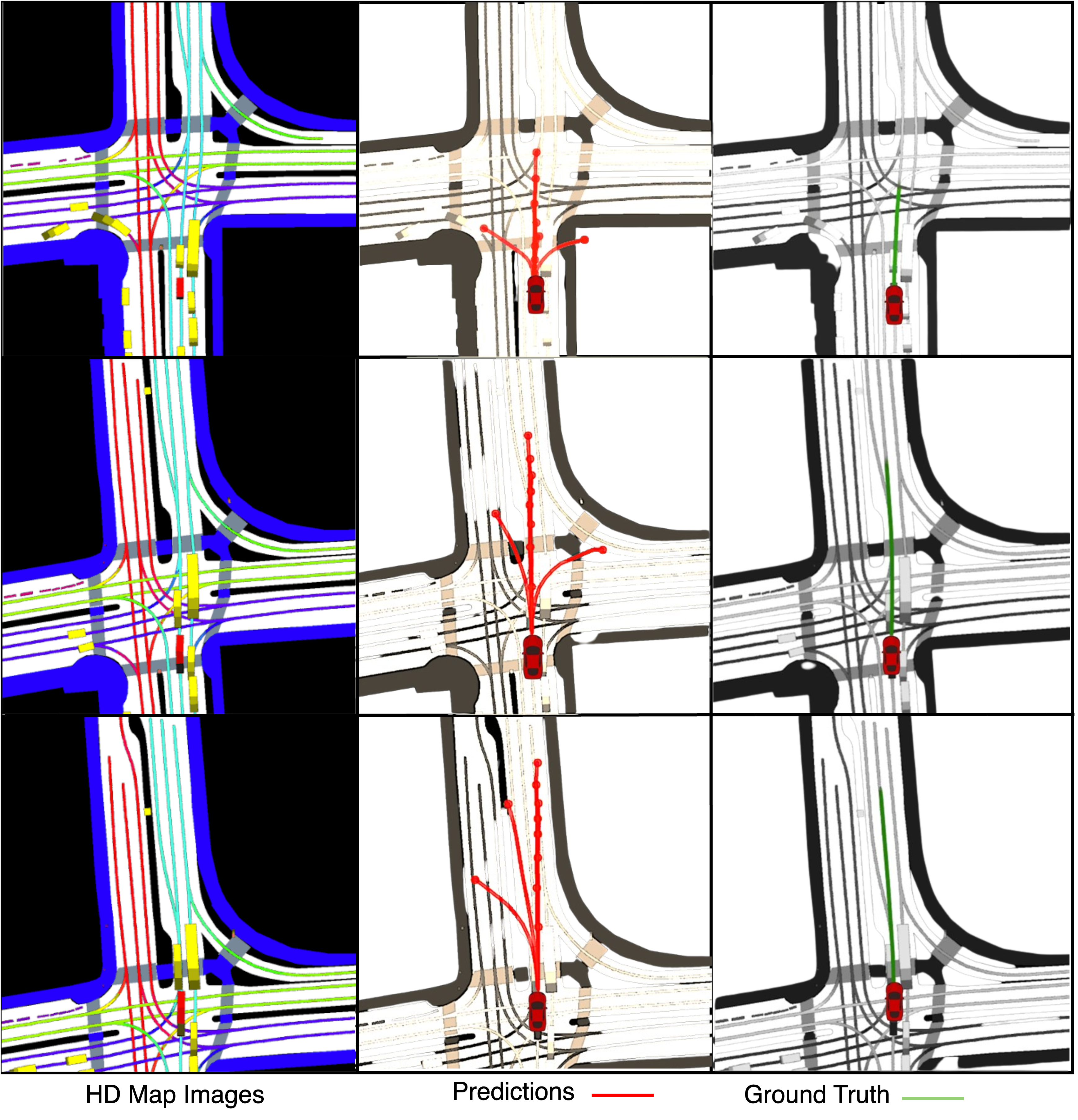}
\caption{\textbf{Qualitative Analysis}: HD Map Integration for Vehicle Trajectory Prediction. The process begins with the feature extractor analyzing these inputs to create image features. These features are then analyzed by a probabilistic network, which evaluates multiple potential trajectories and assigns probabilities to each. On the left, the HD image is displayed; the middle section shows the multimodal trajectory prediction in \textcolor{red}{red}; and on the right, the ground truth trajectory is shown in \textcolor{green}{green}.}
\label{fig:HDresults}
\end{figure*}

As illustrated in Figure \ref{fig:HDresults}, on the left side, we utilize HD map or raster map data from Nuscene. The HD map provides detailed and precise information regarding the environment surrounding the vehicle. These map information  include accurate lane markings, road boundaries, traffic signs, and other objects present in the scene. In the middle, the multi-model trajectory prediction is depicted in \textcolor{red}{red}. This visualization showcases multiple possible trajectories, each representing a potential path that a vehicle might take, considering various factors such as speed and direction. It provides valuable insights into the diverse range of paths that vehicles could follow in different scenarios. Note that,  our model's trajectory predictions are well-represented, as evidenced by the ground truth depicted on the right-hand side in \textcolor{green}{green}. This alignment with ground truth demonstrates the efficacy of our model in accurately forecasting vehicle trajectories.  In Figure \ref{fig:HDresults}, we consider the highest probability to predict the trajectory, but we also take into account other probabilities for different scenarios.

\section{Conclusion}

In this paper, we present a straightforward solution for MapsTP by utilizing them alongside state variables as inputs. We employ a basic CNN architecture like ResNet-50 to extract features from the HD map. Subsequently, we employ an STPN to compute potential trajectories. Our method shows promising results for accurate and efficient trajectory prediction, outperforming current state-of-the-art techniques. The multi-model trajectory prediction allows for the calculation of multiple possible paths in a scene, aiding in better and more precise decision-making.
In future research, we aim to enhance trajectory prediction for vehicles by integrating multiple sensors such as Lidar and radar. Additionally, we plan to incorporate other datasets like Argoverse to further improve the robustness and generalization of our approach.

\section*{Acknowledgments}
%Anonymous
This publication has emanated from research supported in part by a grant from Science Foundation Ireland under Grant number 18/CRT/6049. For the purpose of Open Access, the author has applied a CC BY public copyright license to any Author Accepted Manuscript version arising from this submission.

%%%%%%%%%%%%%%%%%%%%%%%%
\bibliographystyle{apalike}

\bibliography{imvip}

\end{document}